\documentclass[10pt,twocolumn,letterpaper]{article}

\usepackage{cvpr}
\usepackage{times}
\usepackage{epsfig}
\usepackage{graphicx}
\usepackage{amsmath}
\usepackage{amssymb}
\usepackage{float}
\usepackage{subfigure}

\usepackage[breaklinks=true,bookmarks=false]{hyperref}

\cvprfinalcopy % *** Unco1mment this line for the final submission

\def\cvprPaperID{793} % *** Enter the CVPR Paper ID here

\begin{document}

%%%%%%%%% TITLE
%\title{Unsupervised Learning of Long-Term Temporal Representations for Videos}
\title{Unsupervised Learning of Long-Term Motion Dynamics for Videos}

\author{Zelun Luo, Boya Peng, De-An Huang, Alexandre Alahi, Li Fei-Fei\\
Stanford University \\
{\tt\small \{zelunluo,boya,dahuang,alahi,feifeili\}@cs.stanford.edu}
% \and
% % First line of institution2 address\\
}

\maketitle
%\thispagestyle{empty}

%%%%%%%%% ABSTRACT
\begin{abstract}
We present an unsupervised representation learning approach that compactly encodes the motion dependencies in videos. Given a pair of images from a video clip, our framework learns to predict the long-term 3D motions. To reduce the complexity of the learning framework, we propose to describe the motion as a sequence of atomic 3D flows computed with RGB-D modality. We  use a Recurrent Neural Network based Encoder-Decoder framework to predict these sequences of flows. We argue that in order for the decoder to reconstruct these sequences, the encoder must learn a robust video representation that captures long-term motion dependencies and spatial-temporal relations. We demonstrate the effectiveness of our learned temporal representations on  activity classification across multiple modalities and datasets such as NTU RGB+D and  MSR Daily Activity 3D. Our framework is generic to any input modality, \textit{i.e.}, RGB, depth, and RGB-D videos.
%Our learned representation outperforms previous work.
\end{abstract}

%%%%%%%%% BODY TEXT
\section{Introduction}

% Motivation

\begin{figure}[t]
\centering
\includegraphics[width=\linewidth]{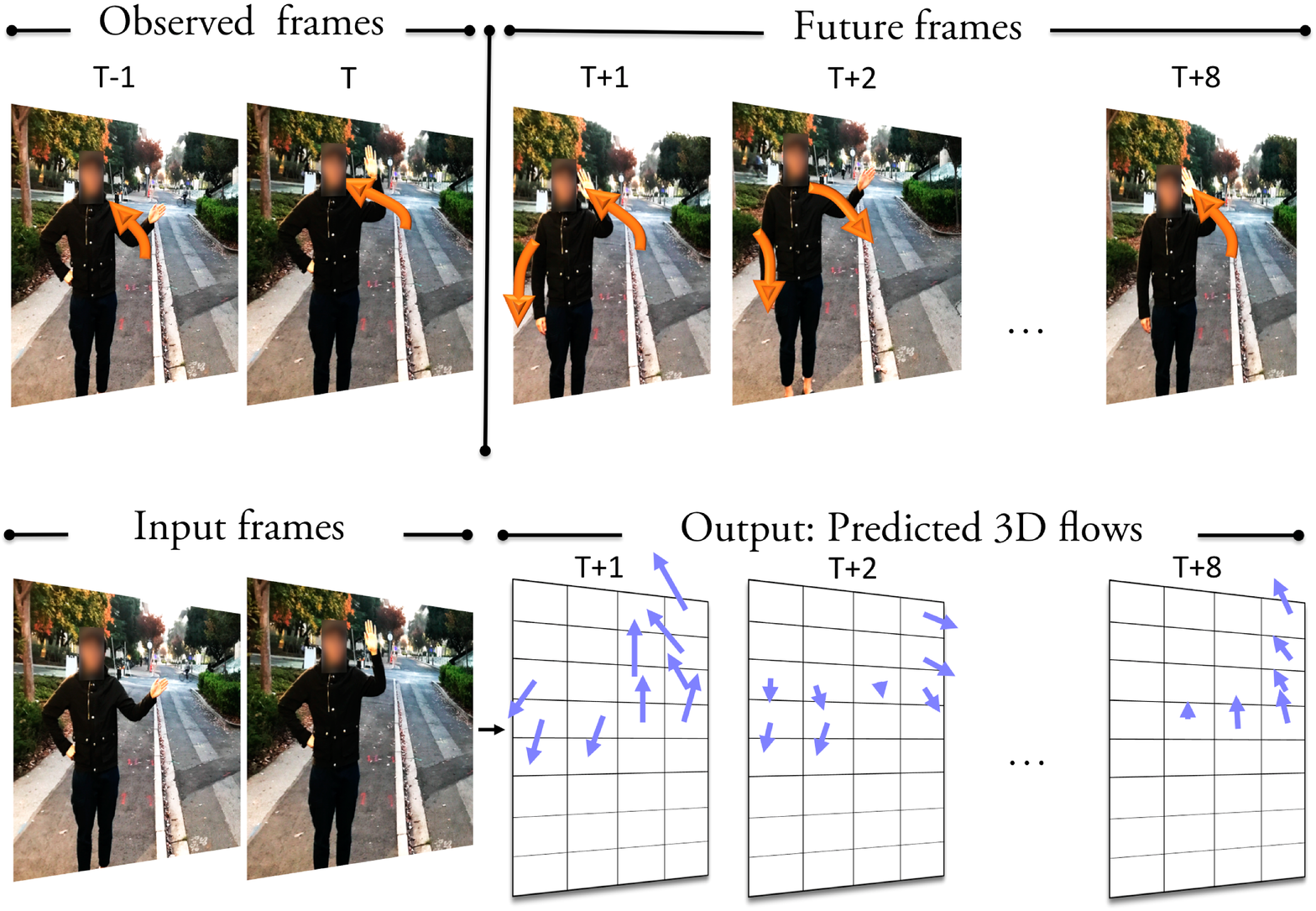}
\caption{
We propose a method that learns a video representation by predicting a sequence of basic motions described as atomic 3D flows. The learned representation is then extracted from this model to recognize activities.
%We propose a self-supervised learning scheme that can learn a feature representation by predicting a long-term motions in videos. This representation is shown to generalize well to various activity recognition tasks.
}
\label{fig:pull}
\end{figure}

% What the pb?
Human activities can often be described as a sequence of basic motions. For instance, common activities like brushing hair or waving a hand can be described as a sequence of successive raising and lowering of the hand. Over the past years, researchers have studied multiple strategies to effectively represent motion dynamics and classify activities in videos \cite{wang2013action,oreifej2013hon4d,wang2015towards}. However, the existing methods suffer from the inability to compactly encode long-term motion dependencies. In this work, we propose to learn a representation that can describe the sequence of motions by learning to predict it. In other words, we are interested in learning a representation that, given a pair of video frames, can predict the sequence of basic motions (see in Figure \ref{fig:pull}). We believe that if the learned representation has encoded enough information to predict the motion, it is discriminative enough to classify activities in videos. Hence, our final goal is to use our learned representation to classify activities in videos.

% Why? Pb statement
To classify activities, we argue that a video representation needs to capture not only the semantics, but also the motion dependencies in a long temporal sequence. Since robust representations exist to extract semantic information \cite{simonyan2014very}, we focus our effort on learning a representation that encodes the sequence of basic motions in consecutive frames. We define basic motions as atomic 3D flows. The atomic 3D flows are computed by quantizing the estimated dense 3D flows in space and time using RGB-D modality. Given a pair of images from a video clip, our framework learns a representation that can predict the sequence of atomic 3D flows.

% What are the benefits of our framework?
Our learning framework is unsupervised, \textit{i.e.}, it does not require human-labeled data.  Not relying on labels has the following benefits. It is not clear how many labels are needed to understand activities in videos. For a single image, millions of labels have been used to surpass human-level accuracy in extracting semantic information \cite{simonyan2014very}. Consequently, we would expect that videos will require several orders of magnitude more labels to learn a representation in a supervised setting. It will be unrealistic to collect  all these labels.

Recently, a stream of unsupervised methods have been proposed to learn temporal structures from videos. These methods are formulated with various objectives - supervision. Some focus on constructing future frames \cite{srivastava2015unsupervised, ranzato2014video}, or enforcing the learned representations to be temporally smooth \cite{zou2012deep}, while others make use of the sequential order of frames sampled from a video \cite{misra2016shuffle, wang2015unsupervised}. Although they show promising results, most of the learned representations still focus heavily on either capturing semantic features  \cite{misra2016shuffle}, or are not discriminative enough for classifying activities as the output supervision is too large and coarse (\textit{e.g.}, frame reconstruction).

%In Section \ref{subsec: unsupervise}, we quantitatively show the limitations of these existing methods.

When learning a representation that predicts motions, the following properties are needed: the output supervision needs to be of i) low dimensionality, ii) easy to parameterize, and iii) discriminative enough for other tasks. We address the first two properties by reducing the dimensionality of the flows through clustering. Then, we address the third property by augmenting the RGB videos with depth modality to reason on 3D motions. By inferring 3D motion as opposed to view-specific 2D optical flow, our model is able to learn an intermediate representation that captures less view-specific spatial-temporal interactions. Compared to 2D dense trajectories \cite{wang2013action},  our 3D motions are of much lower dimensionality. Moreover, we focus on inferring the sequence of basic motions that describes an activity as opposed to tracking keypoints over space and time. We claim that our proposed description of the motion enables our learning framework to predict longer motion dependencies since the complexity of the output space is reduced. In Section \ref{subsec: supervise}, we show quantitatively that our proposed method outperforms previous methods on activity recognition.

The contributions of our work are as follows:
\begin{itemize}
\item[(i)] We propose to use a Recurrent Neural Network based Encoder-Decoder framework to effectively learn a representation that predicts the sequence of basic motions. Whereas existing unsupervised methods describe motion as either a single optical flow \cite{walker2015dense} or 2D dense trajectories \cite{wang2013action}, we propose to describe it as a sequence of atomic 3D flows over a long period of time (Section \ref{sec:method}).
\item[(ii)] We are the first to explore and generalize unsupervised learning methods across different modalities. We study the performance of our unsupervised task - predicting the sequence of basic motions - using various input modalities: RGB $\rightarrow$ motion, depth $\rightarrow$ motion, and RGB-D $\rightarrow$ motion (Section \ref{subsec: unsupervise}).
\item[(iii)] We show the effectiveness of our learned representations on activity recognition tasks across multiple modalities and datasets (Section \ref{subsec: supervise}). At the time of its introduction, our model outperforms state-of-the-art unsupervised methods \cite{misra2016shuffle,srivastava2015unsupervised} across modalities (RGB and depth).
\end{itemize}

\begin{figure*}[t]
\centering
\includegraphics[width=1.0\linewidth]{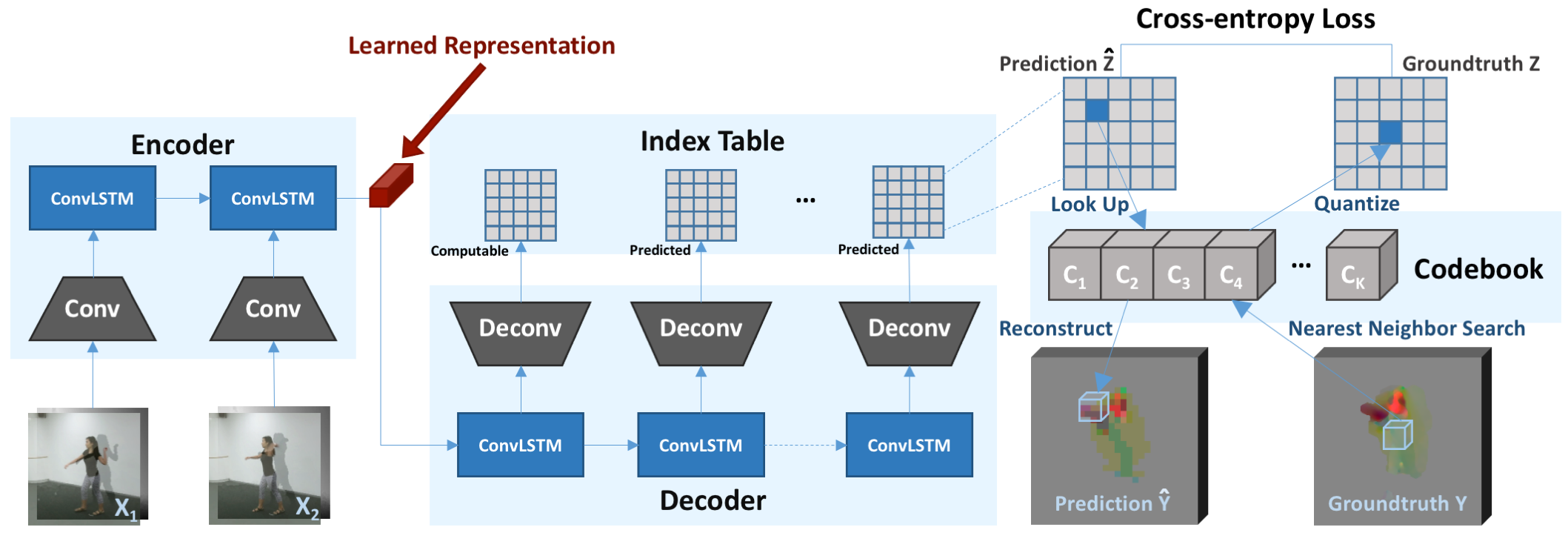}
\caption{Our proposed learning framework based on the LSTM Encoder-Decoder method. During the encoding step, a downsampling network (referred to as ``Conv") extracts a low-dimensionality feature from the input frames. Note that we use a pair of frames as the input to reduce temporal ambiguity. Then, the LTSM learns a temporal representation. This representation is then decoded with the upsampling network (referred to as ``Deconv") to output the atomic 3D flows.}
\label{fig:arch}
\end{figure*}

\section{Related Work}
We first present previous works on unsupervised representation learning for images and videos. Then, we give a brief overview on existing methods that classify activities in multi-modal videos.

\noindent\textbf{Unsupervised Representation Learning.} In the RGB domain, unsupervised learning of visual representations has shown usefulness for various supervised tasks such as pedestrian detection and object detection \cite{doersch2015unsupervised, sermanet2013pedestrian}.
% It has a rich history starting from the original auto-encoders work of Olhausen and Field \cite{olshausen1997sparse}. Deep learning methods like auto-encoders \cite{olshausen1996emergence, thibodeau2014deep, vincent2008extracting}, Deep Boltzmann Machines \cite{salakhutdinov2009deep}, variational methods \cite{kingma2013auto, rezende2014stochastic} and stacked auto-encoders \cite{lee2006efficient, bengio2007greedy} learn representations directly from images and captures mostly semantic features.
To exploit temporal structures, researchers have started focusing on learning visual representations using RGB videos. Early works such as \cite{zou2012deep} focused on inclusion of constraints via video to autoencoder framework. The most common constraint is enforcing learned representations to be temporally smooth \cite{zou2012deep}. More recently, a stream of reconstruction-based models has been proposed. Ranzato et al. \cite{ranzato2014video} proposed a generative model that uses a recurrent neural network to predict the next frame or interpolate between frames. This was extended by Srivastava et al. \cite{srivastava2015unsupervised} where they utilized a LSTM Encoder-Decoder framework to reconstruct current frame or predict future frames. Another line of work \cite{wang2015unsupervised} uses video data to mine patches which belong to the same object to learn representations useful for distinguishing objects. Misra et al. \cite{misra2016shuffle} presented an approach to learn visual representation with an unsupervised sequential verification task, and showed performance gain for supervised tasks like activity recognition and pose estimation. One common problem for the learned representations is that they capture mostly semantic features that we can get from ImageNet or short-range activities, neglecting the temporal features.

\noindent\textbf{RGB-D / depth-Based Activity Recognition.} Techniques for activity recognition in this domain use appearance and motion information in order to reason about non-rigid human deformations activities. Feature-based approaches such as HON4D \cite{oreifej2013hon4d}, HOPC \cite{rahmani2014hopc}, and DCSF \cite{xia2013spatio} capture spatio-temporal features in a temporal grid-like structure. Skeleton-based approaches such as \cite{haque2016viewpoint,rahmani20163d,vemulapalli2014human,wang2012mining,yu2014discriminative} move beyond such sparse grid-like pooling and focus on how to propose good skeletal representations. Haque \etal \cite{haque2016recurrent} proposed an alternative to skeleton representation by using a Recurrent Attention model (RAM). Another stream of work uses probabilistic graphical models such as Hidden Markov Models (HMM) \cite{yang2013discovering}, Conditional Random Fields (CRF) \cite{koppula2013learning} or Latent Dialect Allocation (LDA) \cite{wu2015watch} to capture spatial-temporal structures and learn the relations in activities from RGB-D videos. However, most of these works require a lot of feature engineering and can only model short-range action relations.
State-of-the-art methods \cite{lu2014range, luo2013group} for RGB-D/depth-based activity recognition report human level performance on well-established datasets like MSRDailyActivity3D \cite{li2010action} and CAD-120 \cite{sung2011human}. However, these datasets were often constructed under various constraints, including single-view, single background, or with very few subjects. On the other hand, \cite{Shahroudy_2016_CVPR} shows that there is a big performance gap between human and existing methods on a more challenging dataset \cite{Shahroudy_2016_CVPR}, which contains significantly more subjects, viewpoints, and background information.

\noindent\textbf{RGB-Based Activity Recognition.} The past few years have seen great progress on activity recognition on short clips \cite{lan2015beyond, yue2015beyond,simonyan2014two, wang2013action, wang2013mining}. These works can be roughly divided into two categories. The first category focuses on handcrafted local features and Bag of Visual Words (BoVWs) representation. The most successful example is to extract improved trajectory features \cite{wang2013action} and employ Fisher vector representation \cite{sanchez2013image}. The second category utilizes deep convolutional neural networks (ConvNets) to learn video representations from raw data (\textit{e.g.}, RGB images or optical flow fields) and train a recognition system in an end-to-end manner. The most competitive deep learning model is the deep two-stream ConvNets \cite{wang2015towards} and its successors \cite{WangXW015,Wang0T15}, which combine both semantic features extracted by ConvNets and traditional optical flow that captures motion. However, unlike image classification, the benefit of using deep neural networks over traditional handcrafted features is not very evident. This is potentially because supervised training of deep networks requires a lot of data, whilst the current RGB activity recognition datasets are still too small.

\begin{figure*}[t]
\centering
\includegraphics[width=\linewidth]{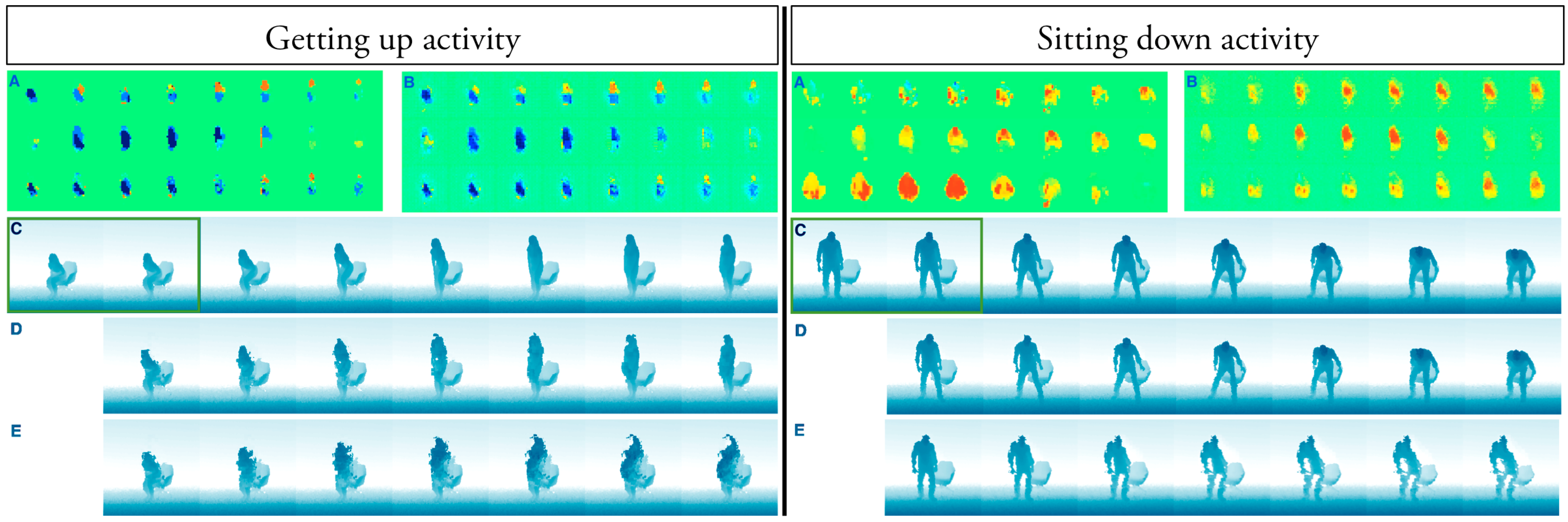}
\caption{Qualitative results on predicting motion: two examples of long-term flow prediction (8 timesteps, 0.8s). The right hand side illustrates the ``Getting up" activity whereas the right side presents the ``Sitting down" activity. A: Ground truth 3D flow. Each row corresponds to flow along x, y, z direction respectively. B: Predicted 3D flows. C: Ground truth depths. The two frames in green boxes are the input. D: Depth reconstructed by adding ground truth depth and predicted flow. E: Depth reconstructed by adding the previous reconstructed depth and predicted flow, except for the first frame, in which case the ground truth depth is used. }
\label{fig:visualize}
\end{figure*}

\section{Method}\label{sec:method}
% Overview
The goal of our method is to learn a representation that predicts the sequence of basic motions, which are defined as atomic 3D flows (described in details in Section \ref{atomic}). The problem is formulated as follows: given a pair of images $\langle \mathbf{X_{1}}, \mathbf{X_{2}} \rangle$, our objective is to predict the sequence of atomic 3D flows over $T$ temporal steps: $\langle \mathbf{\widehat{Y}_{1}}, \mathbf{\widehat{Y}_{2}}, ..., \mathbf{\widehat{Y}_{T}} \rangle$, where $\mathbf{\widehat{Y}_{t}}$ is the atomic 3D flow at time $t$ (see Figure \ref{fig:arch}). Note that $\mathbf{X_{i}} \in \mathbb{R}^{H \times W \times \mathcal{D}}$ and $\mathbf{\widehat{Y}_{t}} \in \mathbb{R}^{H \times W \times 3}$, where $\mathcal{D}$ is the number of input channels, and $H, W$ are the height and width of the video frames respectively. In Section \ref{sec:exp}, we experiment with inputs from three different modalities: RGB only ($\mathcal{D}=3$), depth only ($\mathcal{D}=1$), and RGB-D ($\mathcal{D}=4$).

The learned representation -- the red cuboid in Figure \ref{fig:arch} -- can then be used as a motion feature for activity recognition (as described in Section \ref{sec:activity}). In the remaining of this section, we first present details on how we describe basic motions. Then, we present the learning framework .

\subsection{Sequence of Atomic 3D Flows}\label{atomic}
To effectively predict the sequence of basic motions, we need to describe the motion as a low-dimensional signal such that it is easy to parameterize and is discriminative enough for other tasks such as activity recognition. Inspired by the vector quantization algorithms for image compression \cite{DBLP:dblp_journals/tip/Kim92}, we propose to address the first goal by quantizing the estimated 3D flows in space and time, referred to as atomic 3D flows. We address the discriminative property by inferring a long-term sequence of 3D flows instead of a single 3D flow. With these properties, our learned representation has the ability to capture longer term motion dependencies.

\noindent\textbf{Reasoning in 3D.} Whereas previous unsupervised learning methods model 2D motions in the RGB space \cite{walker2015dense}, we propose to predict motions in 3D. The benefit of using depth information along with RGB input is to overcome difficulties such as variations of texture, illumination, shape, viewpoint, self occlusion, clutter and occlusion. We augment the RGB videos with depth modality and estimate the 3D flows \cite{jaimez2015primal} in order to reduce the level of ambiguities that exist in each independent modality.

\noindent\textbf{Reasoning with sequences.} Previous unsupervised learning methods have modeled motion as either a single optical flow \cite{walker2015dense} or a dense trajectories over multiple frames \cite{wang2013action}. The first approach has the advantage of representing motion with a single fixed size image. However, it only encodes a short range motion. The second approach addresses the long-term motion dependencies but is difficult to efficiently model each keypoint. We propose a third alternative: model the motion as a sequence of flows. Motivated by the recent success of RNN to predict sequence of images \cite{NIPS2014_5346}, we propose to learn to predict the sequence of flows over a long period of time. To ease the prediction of the sequence, we can further transform the flow into a lower dimensionality signal (referred to as atomic flows).

\noindent\textbf{Reasoning with atomic flows.}
Flow prediction can be posed as a regression problem where the loss is squared Euclidean distance between the ground truth flow and predicted flow. Unfortunately, the squared Euclidean distance in pixel space is not a good metric, since it is not stable to small image deformations, and the output space tends to smoothen results
to the mean \cite{ranzato2014video}. Instead, we formulate the flow prediction task as a classification task using $\mathbf{Z}=\mathcal{F}(\mathbf{Y})$, where $\mathbf{Y} \in \mathbb{R}^{H \times W \times 3}$, $\mathbf{Z} \in \mathbb{R}^{h \times w \times K}$, and $\mathcal{F}$ maps each non-overlapping $M \times M$ 3D flow patch in $\mathbf{Y}$ to a probability distribution over $K$ quantized classes (\textit{i.e.}, atomic flows).
More specifically, we assign a soft class label over $K$ quantized codewords for each $M \times M$ flow patch, where $M=H/h=W/w$. After mapping each patch to a probability distribution, we get a probability distribution $\mathbf{Z} \in \mathbb{R}^{h \times w \times K}$ over all patches. We investigated three quantization methods: k-means codebook (similar to \cite{walker2015dense}), uniform codebook, and learnable codebook (initialized with k-means or uniform codebook, and trained end-to-end). We got the best result using uniform codebook and training the codebook end-to-end only leads to minor performance gain. K-means codebook results in inferior performance because the lack of balance causes k-means to produce a poor clustering.

Our uniform quantization is performed as follows: we construct a codebook $\mathbf{C} \in \mathbb{R}^{K \times 3}$ by quantizing bounded 3D flow into equal-sized bins, where we have $\sqrt[3]{K}$ distinct classes along each axes. For each $M \times M$ 3D flow patch, we compute its mean and retrieve its $k$ nearest neighbors (each represents one flow class) from the codebook. Empirically, we find having the number of nearest neighbors $k > 1$ (soft label) yields better performance.
To reconstruct the predicted flow $\mathbf{\widehat{Y}}$ from predicted distribution $\mathbf{\widehat{Z}}$, we replace each codebook distribution as a linear combination of codewords. The parameters are determined empirically such that $K=125$ ($5$ quantized bins across each dimension) and $M=8$.

\subsection{Learning framework}
To learn a representation that encodes the long-term motion dependencies in videos, we cast the learning framework as a sequence-to-sequence problem. We propose to use a Recurrent Neural Network (RNN) based Encoder-Decoder framework to effectively learn these motion dependencies. Given two frames, our proposed RNN predicts the sequence of atomic 3D flows.

Figure \ref{fig:arch} presents an overview of our learning framework, which can be divided into an encoding and decoding steps. During encoding, a downsampling network (referred to as ``Conv") extracts a low-dimensionality feature from the input frames. Then, the LTSM runs through the sequence of extracted features to learn a temporal representation. This representation is then decoded with the upsampling network (``Deconv") to output the atomic 3D flows.

\begin{figure}[t]
\centering
\includegraphics[width=0.4\linewidth]{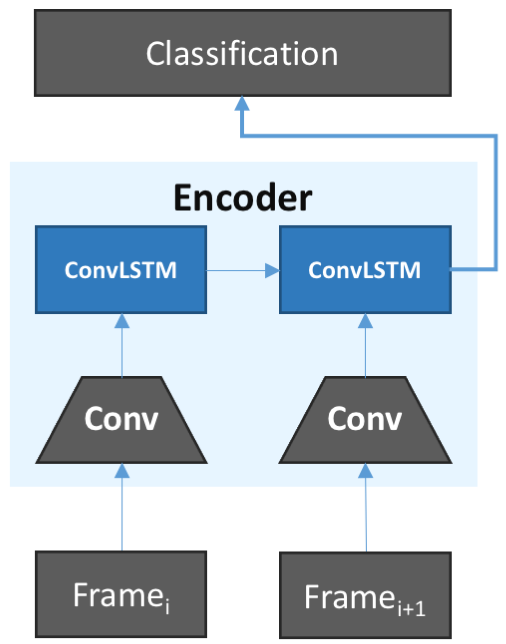}
\caption{Our proposed network architecture for activity recognition. Each pair of video frames is encoded with our learned temporal representation (fixing the weights). Then, a classification layer is trained to infer the activities.}
\label{fig:sup}
\end{figure}

The LSTM Encoder-Decoder framework \cite{NIPS2014_5346} provides a general framework for sequence-to-sequence learning problems, and its ability to capture long-term temporal dependencies makes it a natural choice for this application. However, vanilla LSTMs do not take spatial correlations into consideration. In fact, putting them between the upsampling and downsampling networks leads to much slower convergence speed and significantly worse performance, compared to a single-step flow prediction without LSTMs. To preserve the spatial information in intermediate representations, we use the convolutional LSTM unit \cite{xingjian2015convolutional} that has convolutional structures in both the input-to-state and state-to-state transitions. Here are more details on the downsampling and upsampling networks:

\noindent\textbf{Downsampling Network (``Conv" ).}
We train a Convolutional Neural Network (CNN) to extract high-level features from each input frame. The architecture of our network is similar to the standard VGG-16 network \cite{simonyan2014very} with the following modifications. Our network is fully convolutional, with the first two fully connected layers converted to convolution with the same number of parameters to preserve spatial information. The last softmax layer is replaced by a convolutional layer with a filter of size  $1 \times 1 \times 32$, resulting in a downsampled output of shape $7 \times 7 \times 32$. A batch normalization layer \cite{DBLP:journals/corr/IoffeS15} is added to the output of every convolutional layer. In addition, the number of input channels in the first convolutional layer is adapted according to the modality.

\noindent\textbf{Upsampling Network (``Deconv").} We use an upsampling CNN with fractionally-strided convolution \cite{springenberg2014striving} to perform spatial upsampling and atomic 3D flow prediction. A stack of five fractionally-strided convolutions upsamples each input to the predicted distribution $\mathbf{\widehat{Z}} \in \mathbb{R}^{h \times w \times K}$, where $\mathbf{\widehat{Z}_{ij}}$ represents the unscaled log probabilities over the $(i, j)^{th}$ flow patch.

\begin{figure}[t]
\centering
\includegraphics[width=1.0\linewidth]{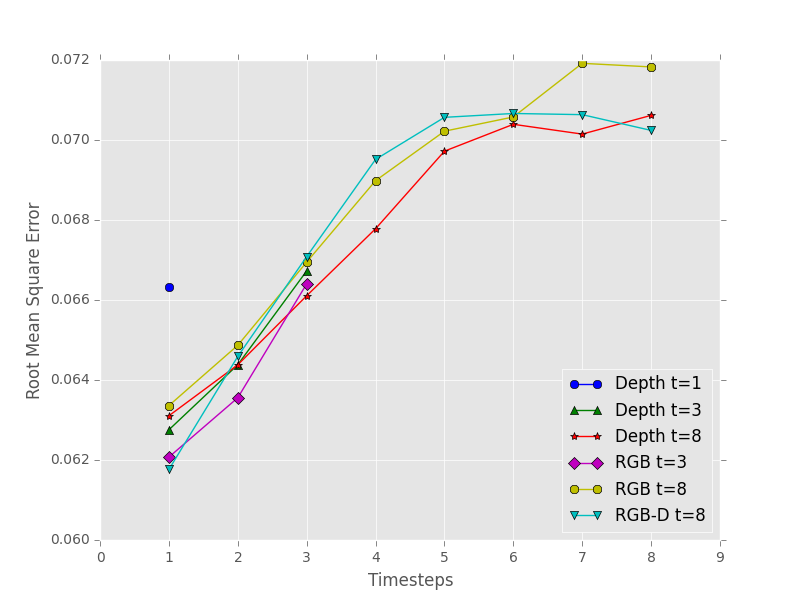}
\caption{Motion prediction error on NTU-RGB+D. We plot the per-pixel root mean square error of estimating the atomic 3D flows with respect to time across different input modalities.}
\label{fig:rmse}
\end{figure}

\subsection{Loss Function}
Finally, we define a loss function that is stable and easy to optimize for motion prediction. As described in section \ref{atomic}, we define the cross-entropy loss between the ground truth distribution $\mathbf{Z}$ over the atomic 3D flow space $\mathbf{C}$ and the predicted distribution $\mathbf{\widehat{Z}}$:

% Previous work on predicting optical flow \cite{walker2015dense} uses the per-pixel regression loss between the ground truth flow $\mathbf{Y}$ and the predicted flow $\mathbf{\widehat{Y}}$ :
% \begin{equation}
% \mathcal{L}_{r}(\mathbf{Y}, \mathbf{\widehat{Y}})=\frac{1}{HWD}
% \sum_{i=1}^{H}\sum_{j=1}^{W}\sum_{k=1}^{3}\lVert \mathbf{Y_{ijk}-\mathbf{\widehat{Y}_{ijk}}} \rVert_{2}
% \label{eq:loss_r}
% \end{equation}
% This loss function is much harder to optimize compared to a more stable classification loss.

\begin{equation}
\mathcal{L}_{ce}(\mathbf{Z}, \mathbf{\widehat{Z}})=-\sum_{i=1}^{H'}\sum_{j=1}^{W'}\sum_{k=1}^{K}\mathbf{w_{k}}\mathbf{Z_{ijk}}\log{\mathbf{\widehat{Z}_{ijk}}}
\label{eq:loss_ce}
\end{equation}
where $\mathbf{w} \in \mathbb{R}^{K}$ is a weighting vector for rebalancing the loss based on the frequency of each atomic flow vectors.

The distribution of atomic 3D flows is strongly biased towards classes with small flow magnitude, as there is little to no motion in the background. Without accounting for this, the loss function is dominated by classes with very small flow magnitudes, causing the model to predict only class 0 which represents no motion. Following the approach in \cite{DBLP:journals/corr/ZhangIE16}, we define the class weight $\mathbf{w}$ as follow:

\begin{equation}
\mathbf{w}\propto\Big((1-\lambda)\mathbf{\widetilde{p}}+ \frac{\lambda}{K}\Big)^{-1}
\qquad\text{and}\qquad
\sum_{k=1}^{K}\mathbf{\widetilde{p}}_{k}\mathbf{w_{k}}=1
\end{equation}

where $\mathbf{\widetilde{p}}$ is the empirical distribution of the codewords in codebook $\mathbf{C}$, and $\lambda$ is the smoothing weight.

% The total loss is defined as
% \begin{equation}
% \mathcal{L}=\mathcal{L}_{ce}(\mathbf{Z}, \mathbf{\widehat{Z}})+\mu\mathcal{L}_{r}(\mathbf{Y}, \mathbf{\widehat{Y}})
% \label{eq:loss_total}
% \end{equation}
% where $\mu$ is a weight factor.

\section{Activity recognition}\label{sec:activity}
The final goal of our learned representation is to classify activities in videos. We use our encoder architecture from unsupervised learning for activity recognition. A final classification layer is added on top of the encoder output to classify activities.

To study the effectiveness of our learned representation, we consider the following three scenarios:

\begin{enumerate}
\item Initialize the weight of our architecture randomly and learn them with the labels available for the supervision task (referred to as ``architecture only'' in Table \ref{table:ablation});
\item Initialize the weights with our learned representation and fine-tune on activity recognition datasets;
\item Keep the pre-trained encoder fixed and only fine-tune the last classification layer.
\end{enumerate}

Note that we don't combine our learned representation with any pre-trained semantic representation (such as the fc7 representation learned on ImageNet \cite{ILSVRC15}). We argue that for our model to learn to predict the basic motions, it needs to understand the semantic content.

We follow the same data sampling strategy described in \cite{simonyan2014two}. During training, a mini-batch of 8 samples is constructed by sampling from 8 training videos, from each of which a pair of consecutive frames is randomly selected. For scenario (i) and (iii), the learning rate is initially set to $10^{-4}$ with a decay rate of $0.96$ every 2000 steps. For scenario (ii), the initial learning rates of encoder and the final classification layer are set to $10^{-5}$ and $10^{-4}$ respectively, with the same decay rate. At test time, we uniformly sample 25 frames from each video and average the scores across the sampled frames to get the class score for the video.

Our presented classification method is intentionally simple to show the strength of our learned representation. Moreover, our method is computationally effective. It runs in real-time since it consists of a forward pass through our encoder. Finally, our learned representation is compact ($7\times7\times32$) enabling implementation on embedded devices.
%-------------
% Table
\begin{table}[t]
\begin{center}
\begin{tabular}[t]{ |c|c|c|c|}
  \hline
  Methods & Depth & RGB \\
  \hline
  \hline
  \hline
  Our architecture only & 37.5 & 34.1  \\
  \hline
  Our method (with 2D motion) & 58.8  & -- \\
%  \hline
%  Our method (no fine-tuning) & 60.2 & running \\
%  \hline
%  Our method (1-step prediction) & 64.7 & -- \\
  \hline
  Our method (3-step prediction) & 62.5 & 54.7  \\
  \hline
    \hline
      \hline
  \textbf{Our method (8-step prediction)} & \textbf{66.2} & 56 \\
  \hline
\end{tabular}
\end{center}
\caption{Quantitative results on activity recognition using the \textbf{NTU-RGB+D} dataset \cite{Shahroudy_2016_CVPR} with the following input modalities: depth and RGB. We report the mean AP in percentage on our ablation study as well as our complete model (in bold). We report the meanAP in percentage.}
\label{table:ablation}
\end{table}
\section{Experiments}\label{sec:exp}
We first present the performance of our unsupervised learning task, \textit{i.e.}, predicting the sequence of motion, using various input modalities including RGB, depth, and RGB-D. Then, we study the effectiveness of our learned representations on classifying activities across multiple modalities and datasets.
%-------------
% Table
\begin{table}[t]
\begin{center}
\begin{tabular}[t]{ |c|c|}
  \hline
  Methods & Depth \\
  \hline
  \hline
  \hline
  HOG \cite{ohn2013joint} & 32.24 \\
  \hline
  Super Normal Vector \cite{yang2014super} & 31.82 \\
  \hline
  HON4D \cite{oreifej2013hon4d} & 30.56 \\
    \hline
    \hline
  Lie Group \cite{vemulapalli2014human} & 50.08 \\
  \hline
  Skeletal Quads \cite{evangelidis2014skeletal} & 38.62 \\
  \hline
  FTP Dynamic Skeletons \cite{hu2015jointly} & 60.23 \\
  \hline
  \hline
  HBRNN-L \cite{du2015hierarchical} & 59.07 \\
  \hline
  2 Layer P-LSTM \cite{Shahroudy_2016_CVPR} & 62.93  \\
    \hline
    \hline
%  Our method (2D motion) & running \\
 % \hline
%  Our method (no fine-tuning) & 59.9 \\
%  \hline
%  Our method (1-step prediction) & 64.7 \\
  \hline
  Shuffle and Learn \cite{misra2016shuffle} & 47.5 \\
  \hline
  \hline
  \hline
  \textbf{Our method (Unsupervised training)} & \textbf{66.2} \\
  \hline
\end{tabular}
\end{center}
\caption{Quantitative results on depth-based activity recognition using the \textbf{NTU-RGB+D} dataset \cite{Shahroudy_2016_CVPR}. The first group (row) presents the state-of-the-art supervised depth-map based method; the second group reports the supervised skeleton-based methods; The third one includes skeleton-based deep learning methods; The fourth is a recently proposed unsupervised method we implemented; The final row presents our complete model. We report the mean AP in percentage.}
\label{table:resultsNTU}
\end{table}
\subsection{Unsupervised Learning of Long-term Motion}
\label{subsec: unsupervise}
\noindent\textbf{Dataset.}
We use the publicly available NTU RGB+D dataset \cite{Shahroudy_2016_CVPR} to train our unsupervised framework. The dataset contains 57K videos for 60 action classes, 40 subjects and 80 viewpoints. We split the 40 subjects into training and testing groups as described in \cite{Shahroudy_2016_CVPR}. Each group consists of 20 subjects where the training and testing sets have 40,320 and 16,560 samples, respectively.

\noindent\textbf{Training details.}
We use a mini-batch of size $16$. The model is trained for 50 epochs with an initial learning rate of $1e^{-4}$ using the Adam optimizer \cite{DBLP:journals/corr/KingmaB14}. We divide the learning rate by 10 whenever validation accuracy stops going up. The network is $L_{2}$ regularized with a weight decay of $5e^{-4}$. For classification, we use a smoothing $\lambda=0.5$.

\noindent\textbf{Evaluation.} We measure the root mean square error (RMSE) between the ground truth flow $\mathbf{Y}$ and the predicted flow $\mathbf{\widehat{Y}}$. F1 score is used to measure the classification error between the ground truth index table $\mathbf{Z}$ and the predicted index table $\mathbf{\widehat{Z}}$.

\noindent\textbf{Results.} In Figure \ref{fig:rmse}, we plot the prediction error with respect to different input modalities (RGB, depth, RGB-D) and prediction time (3 and 8 timesteps). We also report the prediction error of using a single input frame to predict the next frame similar to \cite{walker2015dense} (blue dot). The error is intuitively the highest since there are ambiguities when reasoning with a single input image.
Interestingly, all input modalities perform very similarly when predicting 8 timesteps. The RGB modality is quite competitive to the other two modalites although the 3D information is not measured. When all 4 channels are used, \textit{i.e.}, RGB-D input, the performance is still similar to using the other modality. The overall error linearly increases with the first 4 frames and stabilizes for the final 4 frames.
All methods that predict only the next 3 frames have similar prediction errors compared to the ones that predict a longer sequence. Consequently, our model has enough capacity to learn a harder problem, \textit{i.e.}, predicting long sequences.
In Figure \ref{fig:visualize}, we qualitatively show the prediction output using depth modality. We illustrate the results by reconstructing the input frame (depth image) from the predicted flows. Our method has not been trained to accurately reconstruct the signal. Nevertheless, the reconstructed signals convey the accuracy of the prediction.

\begin{figure}[t]
\centering
\includegraphics[width=\linewidth]{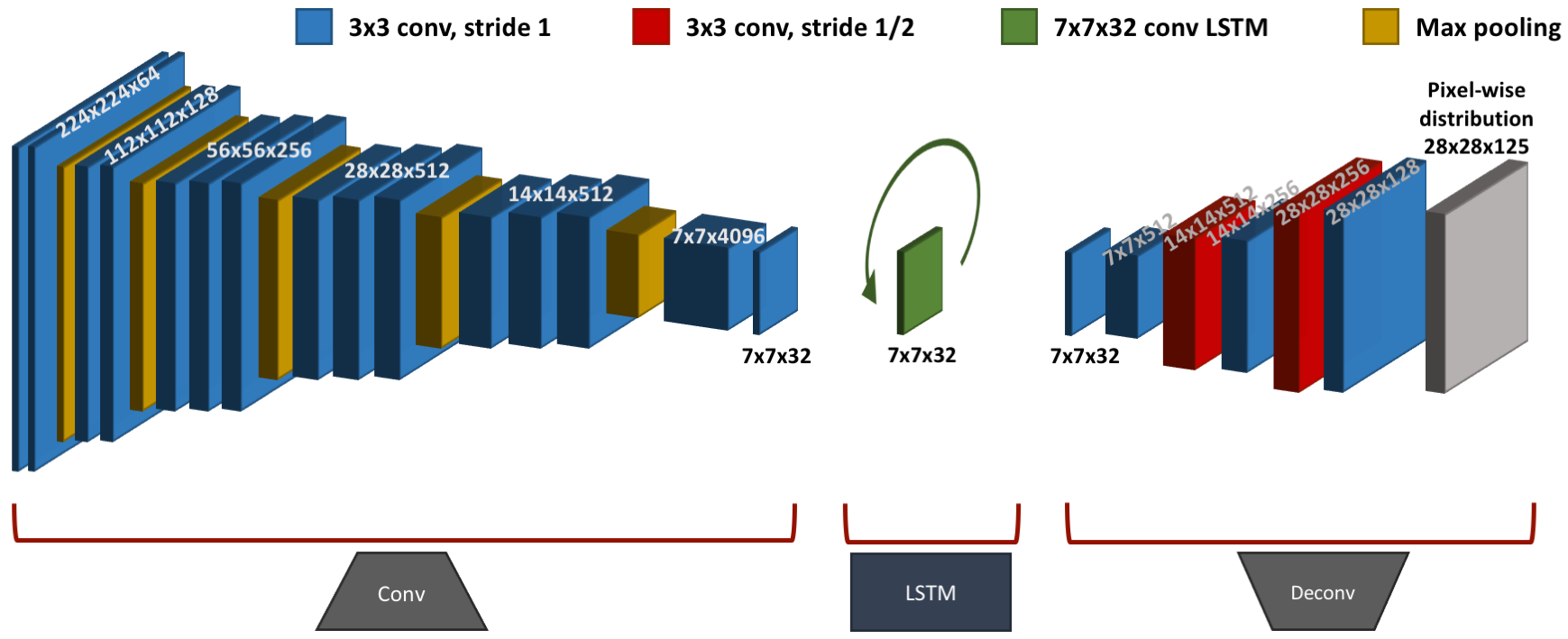}
\caption{The detailed architecture of our model. The Conv shows the architecture of our downsampling network; LSTM represents the encoder-decoder framework; Deconv shows the architecture of our upsampling network.}
\label{fig:detailed}
\end{figure}

\subsection{Activity Recognition}
\label{subsec: supervise}
We compare our activity recognition performance with state-of-the-art supervised methods for each modality. In addition, we perform the ablation studies for our unsupervised methods and compare with the a  recently-proposed unsupervised method.

\noindent\textbf{Our method with 2D motion}. Instead of predicting 3D motion, we predict 2D motion in the form of quantized 2D optical flow.

\noindent\textbf{Our method with 3-step prediction}. We predict motions for the next three frames. Note that our proposed method uses 8-step prediction.

\noindent\textbf{Shuffle and Learn \cite{misra2016shuffle}.} Given a tuple of three frames extracted from a video, the model predicts whether the three frames are in the correct temporal order or not. We implemented the above model using TensorFlow and trained on the NTU RGB-D dataset for the sequential verification task, following the same data sampling techniques and unsupervised training strategies as specified in \cite{misra2016shuffle}.

% In particular, we compare with the following three recently-proposed unsupervised methods in the RGB domain:\\
% \textbf{- Shuffle and Learn \cite{misra2016shuffle}.} Given a tuple of three frames extracted from a video, the model predicts whether the three frames are in the correct temporal order or not. \\
% \textbf{- VGAN \cite{vondrick2016generating}.} Generative adversarial network for video with a spatio-temporal convolutional architecture that generates tiny videos. \\
% \textbf{- Unsupervised LSTMs \cite{srivastava2015unsupervised}.} The model uses an encoder LSTM to map an input sequence into a fixed length representation, which is then decoded to either reconstruct the input sequence or predict the future sequence.\\

\subsubsection{Depth-based Activity Recognition}
\textbf{Dataset.} We train and test our depth-based activity recognition model on two datasets: NTU-RGB+D and MSRDailyActivity3D \cite{li2010action}. For NTU-RGB+D, we follow the cross-subject split as described in \cite{Shahroudy_2016_CVPR}. The MSRDailyActivity3D dataset contains 16 activities performed by 10 subjects. We follow the same leave-one-out training-testing split as in \cite{koperski20143d}. We intentionally use this extra MSRDailyActivity3D dataset that is different from the one we use for unsupervised training to show the effectiveness of our learned representation in new domains (different viewpoints and activities).

\noindent\textbf{Results on NTU-RGB+D.} Table \ref{table:resultsNTU} shows classification accuracy on the NTU-RGB+D dataset. The first group of methods use depth maps as inputs, while the second and the third use skeleton features. Methods in the third group are deep-learning based models. Our proposed method outperforms the state-of-the-art supervised methods. We use our learned representation that predicts the next 8 frames without fine-tuning it on the classification task. Interestingly, fine-tuning the weights of our encoder did not give a boost in performance.

\begin{figure}[t]
\centering
\includegraphics[width=1.0\linewidth]{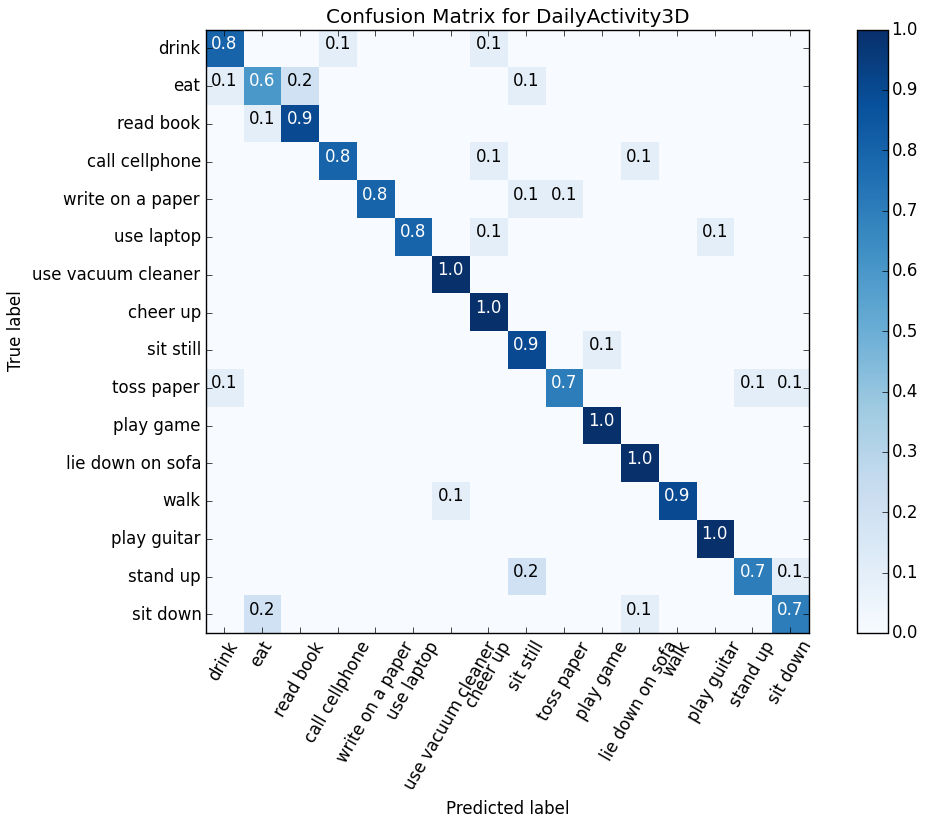}
\caption{The confusion matrix for action recognition on \textbf{MSRDailyActivity3D} dataset \cite{li2010action}. Activities with large motions are better classified than the ones with fine-grained motion.}
\label{fig:confusion2}
\end{figure}

%-------------
% Table
\begin{table}[h!]
\begin{center}
\begin{tabular}[t]{ |c|c|c|c|}
  \hline
  Methods & Depth \\
  \hline
  \hline
  \hline
  \textit{Dynamic Temporal Warping} \cite{muller2006motion} & 54.0 \\
  \hline
  \textit{Actionlet Ensemble} \cite{wang2012mining} & 85.8 \\
  \hline
  \textit{HON4D} \cite{oreifej2013hon4d} & 85 \\
  \hline
  3D Trajectories \cite{koperski20143d} & 72 \\
    \hline
    \hline
  \textbf{Our method (Unsupervised training)} & \textbf{86.9} \\
  \hline
\end{tabular}
\end{center}
\caption{Quantitative results on activity recognition using the \textbf{MSRDailyActivity3D} dataset \cite{li2010action}. Methods in italic require full skeleton detection. Our method has learned a video representation from a different dataset and has not fine-tuned on this dataset. We report the meanAP in percentage.}
\label{table:resultsMSR}
\end{table}

% \begin{figure*}[ht]
% \centering
% \includegraphics[width=0.88\linewidth]{confusion1.png}
% \caption{The confusion matrix for action recognition on NTU dataset \cite{Shahroudy_2016_CVPR}. Activities with large motions are better classified than the ones with fine-grained motion.}
% \label{fig:confusion1}
% \end{figure*}

\noindent\textbf{Ablation study on NTU-RGB+D.} In Table \ref{table:ablation}, we present more insights on our design choices. We first show that by using our encoder architecture without pre-training it to predict the motion (referred to as ``our architecture only''), the classification accuracy (mean AP) is the lowest. We then show that modeling 3D motion instead of 2D motion positively impacts the performance. Finally, we report the results when shorter sequences (3-step prediction) are encoded during our unsupervised training. Increasing the sequence length to 8 time-steps increases the classification accuracy. The discrimination power of our representation is increased by encoding longer-term dependencies. For the sake of completeness, we also fine-tune our activity recognition model using RGB videos from the NTU RGB-D dataset. We notice that the results are comparable to depth-based activity recognition and follow the same trend for ablation studies (\textit{i.e.}, predicting longer motion in 3D yields better performance).

\noindent\textbf{Results on MSRDailyActivity3D.}
Table \ref{table:resultsMSR} presents classification accuracy on the MSRDailyActivity3D dataset \cite{li2010action} and Figure \ref{fig:confusion2} its confusion matrix. Methods in italic require skeleton detection, while the fourth one makes use of dense 3D trajectories. Note that our unsupervised learning task -- predicting the basic motions -- has not been trained on these activities and viewpoints. Nevertheless, we outperform previous work specially the method based on the 3D trajectories by a large margin (+$15\%$). Our compact representation of the 3D motion is more discriminative than the existing representation for 3D trajectories \cite{wang2013action}.

\subsubsection{RGB-based Activity Recognition}
\noindent\textbf{Dataset.} We train and test our RGB-based activity recognition model on the UCF-101 dataset \cite{soomro2012ucf101} to compare with state-of-the-art unsupervised methods \cite{misra2016shuffle,vondrick2016generating} in this domain. The dataset contains 13,320 videos with an average length of 6.2 seconds and 101 different activity categories. We follow the same training and testing protocol as suggested in \cite{simonyan2014two}. However, note that we are not training the unsupervised task on the UCF-101 dataset. Instead, the model is pretrained on the RGB videos from NTU-RGB+D dataset. We want to study the capacity of our learned representation to be used across domains and activities.

\noindent\textbf{Results on UCF-101.} Table \ref{table:resultsUCF} shows classification accuracy for RGB-based activity recognition methods on the UCF-101 dataset. By initializing the weights of our supervised model with the learned representation, our model (\textit{i.e.}, our method w/o semantics) outperforms two recent unsupervised video representation learning approaches \cite{misra2016shuffle,vondrick2016generating}. Note that although the unsupervised LSTM \cite{srivastava2015unsupervised} method outperforms all other methods, it uses a ConvNet pretrained on ImageNet for semantic feature extraction, whilst the other methods do not make use of extra semantic information. To compare with \cite{srivastava2015unsupervised}, we use a VGG-16 network pretrained on ImageNet to extract semantic features (\textit{i.e.}, fc7 feature) from input images, and add a softmax layer on top of it. We combine the softmax score from our model with the semantic softmax score by late fusion.

% Table
\begin{table}[t]
\begin{center}
\begin{tabular}[t]{ |c|c|c|}
  \hline
  Methods & RGB \\
  \hline
  \hline
  \hline
  S: Deep two stream \cite{wang2015towards} & 91.4 \\
  \hline
  \hline
  \hline
  U: Shuffle and Learn \cite{misra2016shuffle} & 50.2 \\
  \hline
  U: VGAN \cite{vondrick2016generating} & 52.1 \\
  \hline
%   U: Our method (Dense trajectory) & \\
%   \hline
  \textbf{U: Our method (w/o semantics)} & \textbf{53.0} \\
  \hline
  \hline
  \hline
  U: Unsupervised LSTMs \cite{srivastava2015unsupervised} & 75.8 \\
    \hline
  \textbf{U: Our method (w/ semantics)} & \textbf{79.3} \\
  \hline
\end{tabular}
\end{center}
\caption{Quantitative results on activity recognition using the \textbf{UCF-101} dataset \cite{soomro2012ucf101}. The first group presents the state-of-the-art supervised (S) method; the second group reports unsupervised (U) methods without using ImageNet semantics; the third shows unsupervised (U) methods with ImageNet semantics. We report the meanAP in percentage.}
\label{table:resultsUCF}
\end{table}

\section{Conclusions}
We have presented a general framework to learn long-term temporal representations for videos across different modalities. By using our proposed sequence of atomic 3D flows as supervision, we can train our model on a large number of unlabeled videos. We show that our learned representation is effective and discriminative enough for classifying actions as we achieve state-of-the-art activity recognition performance on two well-established RGB-D datasets.
For future work, we aim to explore the performance of our method on RGB based datasets such as ActivityNet or other supervised tasks beyond activity recognition. We want to use other free labels from videos such as predicting 3D scenes interactions from RGB frames. We also want to come up with a compact representation for dense trajectory, which can effectively reduce background motions in many existing datasets.

\noindent\textbf{Acknowledgement.} We would like to start by thanking our sponsors: Stanford Computer Science Department, Intel, ONR MURI, and Stanford Program in AI-assisted Care (PAC). Next, we specially thank Juan Carlos Niebles, Serena Yeung, Kenji Hata, Yuliang Zou, and Lyne Tchapmi for their helpful feedback. Finally, we thank all members of the Stanford Vision Lab and Stanford Computational Vision and Geometry Lab for their useful comments and discussions.

\newpage
{\small
\bibliographystyle{ieee}
\bibliography{egbib}
}

\end{document}